\documentclass[10pt,twocolumn,letterpaper]{article}

\usepackage{cvpr}
\usepackage{times}
\usepackage{epsfig}
\usepackage{graphicx}
\usepackage{amsmath}
\usepackage{amssymb}
\usepackage{subcaption}
\usepackage[T1]{fontenc}
\usepackage{threeparttable}

\usepackage[breaklinks=true,bookmarks=false]{hyperref}

\cvprfinalcopy 

\def\cvprPaperID{****} 

\begin{document}

\title{End-to-End Rubbing Restoration Using Generative Adversarial Networks}

\author{Gongbo Sun\\
	Beijing National Day School\\
	Beijing, China\\
\and
Zijie Zheng $^{\ast}$  \\
Beijing National Day School\\
Beijing, China\\
\and
Ming Zhang \thanks{Corresponding author.}\\
School of Computer Science, \\
Peking University\\
mzhang\_cs@pku.edu.cn\\
}

\maketitle

\begin{abstract}
	Rubbing restorations are significant for preserving world cultural history. In this paper, we propose the RubbingGAN model for restoring incomplete rubbing characters. Specifically, we collect characters from the Zhang Menglong Bei and build up the first rubbing restoration dataset. We design the first generative adversarial network for rubbing restoration.\footnote{The dataset and codes are available at \url{https://github.com/qingfengtommy/RubbingGAN}} Based on the dataset we collect, we apply the RubbingGAN to learn the Zhang Menglong Bei font style and restore the characters. The results of experiments show that RubbingGAN can repair both slightly and severely incomplete rubbing characters fast and effectively.
\end{abstract}

\section{Introduction}

\begin{figure}[h!]
	
	\centering
	
	\begin{subfigure}[t]{2.5cm}              
		\includegraphics[width=2.5cm]{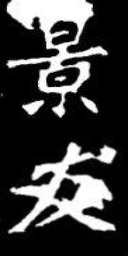}
		\caption{Slightly incomplete Rubbing}
		\label{fig:1left}
	\end{subfigure}
	\quad
	\begin{subfigure}[t]{2.5cm}               
		\includegraphics[width=2.5cm]{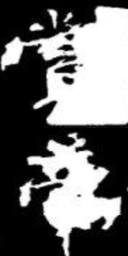}
		\caption{Severely incomplete Rubbing}
		\label{fig:1right}
	\end{subfigure}
	
	\caption{Rubbing characters from ZhangMenglong Bei}
	\label{fig:1}
\end{figure}

Rubbings are ink-on-paper copies of engraved or cast inscriptions and designs on what are mostly cultural objects of metal, stone, and other firm substances \cite{ref1}. The rubbing images are shown in Figure \ref{fig:1}.

The history of rubbings is about 1400 years and rubbings play an important role in providing supportive evidence and clues for cultural relics restoration. However, due to the weathering process in a long time, the characters left on the rubbings usually are incomplete, as shown in Figure~\ref{fig:1right}. Therefor, repairing rubbings is significant for both the Asian history preservation and the world culture inheritance. In traditional rubbings restoration, artists need to confirm the acidity first, and then based on the breakage level to determine corresponding methods. For instance, Li and his colleagues repair Xuan Mi Ta Bei by using a method called bracket mounting (tuo biao fa) \cite{ref2}. However, the traditional rubbing restoration method requires a lot of efforts from artists, which is time consuming and impossible for large-scale artworks repairing.
\par 
In contrast to the traditional manual way of repairing cultural relics, the computer scientists have proposed a digital method called inpainting \cite{ref3}. Digital inpainting is a way of reconstructing missing or damaged part in the digital images or videos, which can be divided into three categories: structural inpainting, textural inpainting, and combined structural and textural inpainting \cite{ref4}. Even though the current inpainting methods can achieve delightful performance in repairing a number of relics such as images \cite{ref5,ref6, ref7}, they cannot be directly applied in rubbing restoration. There are two possible reasons: 1) the structure and texture of the Asian characters on the rubbings are complex and diverse. 2) some Asian characters are similar and there is a slight difference between them.
\par
Another field relevant to rubbing restoration is the font style transfer problem. The font style transfer or the character font generation is a problem which developed from the image style transfer question. Recently, there are some dedicated works on Chinese fonts generation. The most representative work is zi2zi \cite{ref17}.
\par
However, methods in traditional Chinese font style transfer cannot be applied in the rubbing restoration task directly. First, the main objective for rubbing restoration is to restore the incomplete characters, which indicates that we do not know which character needed to generate, and have to recognize it firstly. Second, characters in the datasets used in Chinese font style transfer are usually digital designed characters or just the standard printed fonts, such as KaiTi\_GB2312. The digital or printed font have already be processed elaborately during the designing period of font libraries. And it is relatively easy to transfer the style between two different domain of printed fonts due to the consistency and continuity of the font style. However, in the rubbing restoration question, the characters from the dataset are more aesthetic and unique. The difficulties of the rubbing restoration task include repairing the rubbing characters and recognizing the aesthetic and diverse font style in the rubbings. In other words, we must use the characters which come from the rubbings to realize the job of repairing the incomplete characters and preserving the correspond rubbing style.
\par
In this paper, we propose a method named as Rubbing Generative Adversarial Network~(RubbingGAN) that can restore the incomplete rubbing characters. Specifically, we collect the characters from a rubbing called Zhang Menglong Bei to the dataset. Zhang Menglong Bei has a very high artistic value in the field of calligraphy and has a number of incomplete characters to repair \cite{ref18, ref19}. 
Based on the dataset we collect, we apply the RubbingGAN to restore the characters. 
\par
In short, our main contributions are summarized as follows.
\begin{itemize}
	\item We collect the first rubbing restoration dataset of a Chinese rubbing called Zhang Menglong Bei, including 337 complete characters as the training set, 75 complete characters as the testing set, and 34 incomplete characters as the restoration set. 
	\item We design a RubbingGAN to synthesize the rubbing fonts and repair the incomplete characters based on the training set. As far as we know, it is the first machine learning model in restoring the Chinese rubbing.
	\item The experiments corroborate that the RubbingGAN can learn the rubbing fonts in Zhang Menglong Bei, and can repair both the slightly and severely incomplete characters.
\end{itemize}
\par



\section{Related works}

For the Rubbing restoration problem, we find out that the recognition for the incomplete characters is not the most difficult task. The main difficulty in rubbing restoration is to preserve the aesthetic value of the artworks. Therefor, we mainly focus on artistic style transferring and we regard the rubbing restoration as the image-to-image translation problem. The related works of image-to-image translation are discussed in the below.

\subsection{Generative Adversarial Networks}

The generative adversarial networks is proposed by Ian J. Goodfellow and his colleagues in 2014 \cite{ref39}. The main objective of the GAN is a min-max game. The generator try to deceive the discriminator by generating the high similarity pictures with the source datasets. On the other hand, the discriminator try to recognize the fake and real images. With the training, generator can generate high similarity images with the original input source. 
Based on the Ian J. Goodfellow's work, the M. Mirza and S. Osindero propose a method that directs the data generation by adding the label information into the model \cite{ref40}. And the DCGAN first introduced a deep convolutional network to improve the image quality \cite{ref29}. 
However, the GAN also has some problems. First, the GAN may encounters the mode collapses, which due to the unbalance ability between the generator and the discriminator. Second, the GAN can not converge easily in the training period.
Therefor, there are some works try to solve these problems. For example, the researchers propose WGAN which using the Wasserstein  distance to train the GAN model. And based on the WGAN \cite{ref42}, researchers propose the BEGAN model that aims to match auto-encoder loss distribution using a loss derived from the Wasserstein distance.

\subsection{Image-to-Image translation}

Image-to-image translation is a task of translating one possible representation of a scene into another, given sufficient training data \cite{ref20}. The representative work of translation is a conditional GAN-based model called pix2pix. It performs image-to-image translation on paired data. It uses a "U-Net" \cite{ref43} based architecture as its generator and uses a convolutional "PatchGAN" auto encoder decoder as its discriminator \cite{ref44}. In our work, the generation of rubbing style characters can be regarded as a kind of image-to-image translation. The limitation of the pix2pix model is only supporting the paired datasets \cite{ref20}. Therefor, the researchers propose the model such as CylcleGAN that achieving the cross-domain style transfer \cite{ref34}. However, the model like CylcleGAN can not apply into the rubbing restoration task. Because the precondition of the CycleGAN is that we have the enough data from the different domain. However, this is impractical in the rubbing restoration task. Because there are huge amounts of artworks only have the characters around hundred level, which is not enough for training the CycleGAN model.

\subsection{Character Fonts Generation}

There are a plenty of relative works on neural style transfer or on the character generation. Although the works on the Chinese characters generation are fewer than the English characters, there are some pioneer researches in this field. It has been studied since the beginning of the digital age. The problem can be formulated either as a personal handwritten characters generation problem \cite{ref15}, or an automatic printed font generation problem \cite{ref16}. In the early study of glyph synthesis, researchers focused on the geometric modeling, which relied on the hierarchical representation of simple strokes \cite{ref22,ref23,ref24,ref33,ref34}. They decompose the characters into a hierarchy of reusable components and then to compose the target character in the personalized style. In fact, these methods only focuses on the local characters' features instead of the whole style, in other words, they are limited to particular glyph topology. With the development and popularity of deep neural networks, the topic of style transfer is frequently discussed. Researchers have designed new models which could add the artistic style to image content \cite{ref9, ref10, ref11} successfully. And there are some works used the idea of neural style transfer to do the character generation problem \cite{ref11, ref12, ref13}. This topic gradually evolves to image-to-image translation problem \cite{ref17, ref30, ref37, ref38}. The image-to-image translation problem focuses on converting objects from one domain to another domain. However, most existing methods require an impractical large-size dataset to train the network. For example, \cite{ref17,ref30,ref31,ref32} require 3000 paired characters which costs huge labor resources. In \cite{ref35}, researchers reduce the paired data number to 775 samples. However, this quantities still exceed the total number of the characters in most rubbings. And the dataset they adopt is either font-rendered character images or brush-written character images. Compared with these character images, the rubbing images are more irregular and incomplete. To the best of our knowledge, there is no available collection of rubbing dataset and there is no existing paper using rubbing dataset to conduct experiment.


\section{Dataset}
In this section, We first briefly introduce ZhangMenglong Bei from an artistic perspective. Then we describe the method for data collection and provide the basic information of our collected dataset.


\subsection{A brief introduction of ZhangMenglong Bei}
As we have mentioned above, the existed dataset of Chinese font style are usually generated by the computer, such as KaiTi\_GB2312. Comparing with the traditional rubbing font style characters, these standard printed font style have a very short history and lack of the value for preserving traditional culture. In our task, we care about characters that written on rubbings in ancient ages. Along the history of Chinese calligraphy, there are a huge number of excellent art works that were created during different dynasties. The categories of traditional calligraphy can be divided into five main styles following the chronological order of appearance: Seal Script, Clerical Script, Cursive Script, Semi-Cursive Script and Standard Script (Kai Ti). Between the adjacent two styles, there is a transition style that assimilates the previous core idea and guides the after font style development. The Wei Bei is a famous transitional calligraphy system linking the past and the future. Wei Bei calligraphy refers to the Chinese characters engraved on the tablet during the Wei, Jin, and the Northern and Southern Dynasties. Wei Bei calligraphy has a great influence on the formation of Standard Script in the Later Sui and Tang dynasties. That is to say, Wei Bei calligraphy can be regarded as a predecessor of Standard Script, which has with a high artistic, cultural and research value \cite{ref41}. However, due to the long history, most Wei Bei tablets are usually seriously damaged. A number of characters in each Wei Bei are severely incomplete. Thus, it is necessary to design a method to learn the font style of Wei Bei and restore the incomplete characters. Zhang Menglong Bei is the most prestigious work in the Northern Wei Dynasty and receives high praises from famous calligraphers such as Kang Youwei, Yang Shoujin, Qi gong.  


\subsection{Dataset collection}
\begin{figure}[ht]
	\centering
	
	\begin{subfigure}[2-1]{2.4cm}               
		\includegraphics[width=2.4cm]{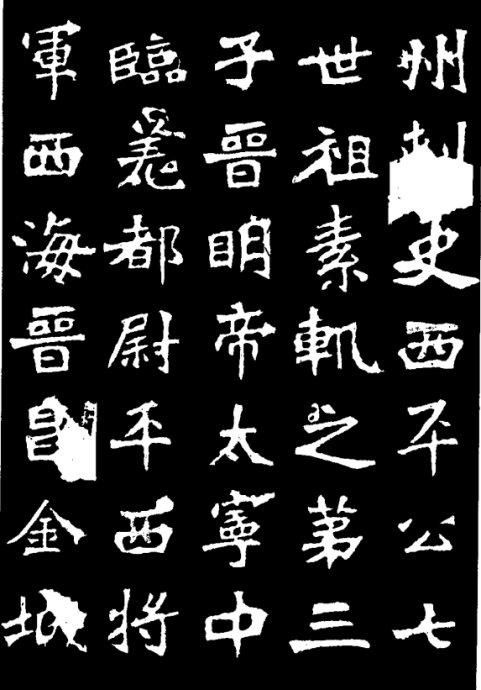}
	\end{subfigure}
	\hfill
	\begin{subfigure}[2-2]{2.4cm}
		\includegraphics[width=2.4cm]{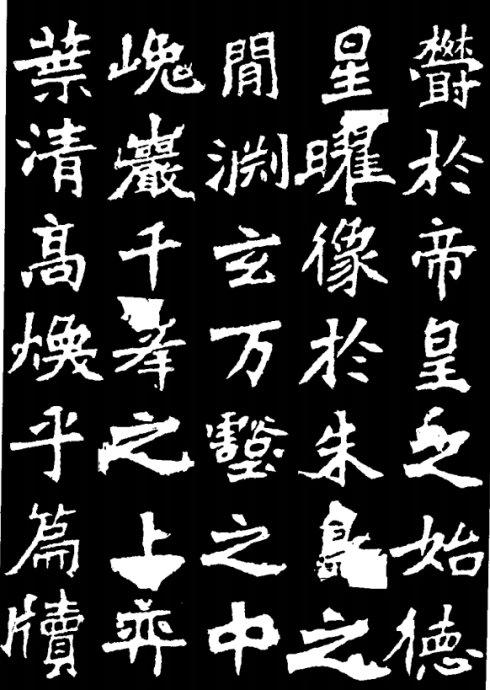}
	\end{subfigure}
	\hfill
	\begin{subfigure}[2-3]{2.4cm}
		\includegraphics[width=2.4cm]{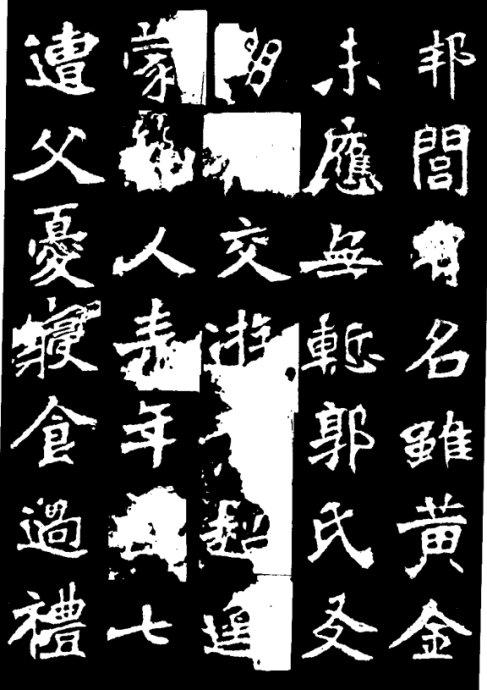}
	\end{subfigure}
	\caption{The images of ZhangMenglong Bei downloaded from the web. 
	}
	\label{fig:2}
\end{figure}

ZhangMenglong Bei has various versions online, and the version we used to extract characters is from the website. We use python to write a web crawler that automatically downloading Zhang Menglong Bei, which is shown in Figure \ref{fig:2}. Then, we can find that Zhang Menglong Bei consist of both the complete characters and the incomplete characters. 
\par
\begin{figure}[h]
	\centering
	\begin{subfigure}{3.8cm}
		\includegraphics[width=3.8cm]{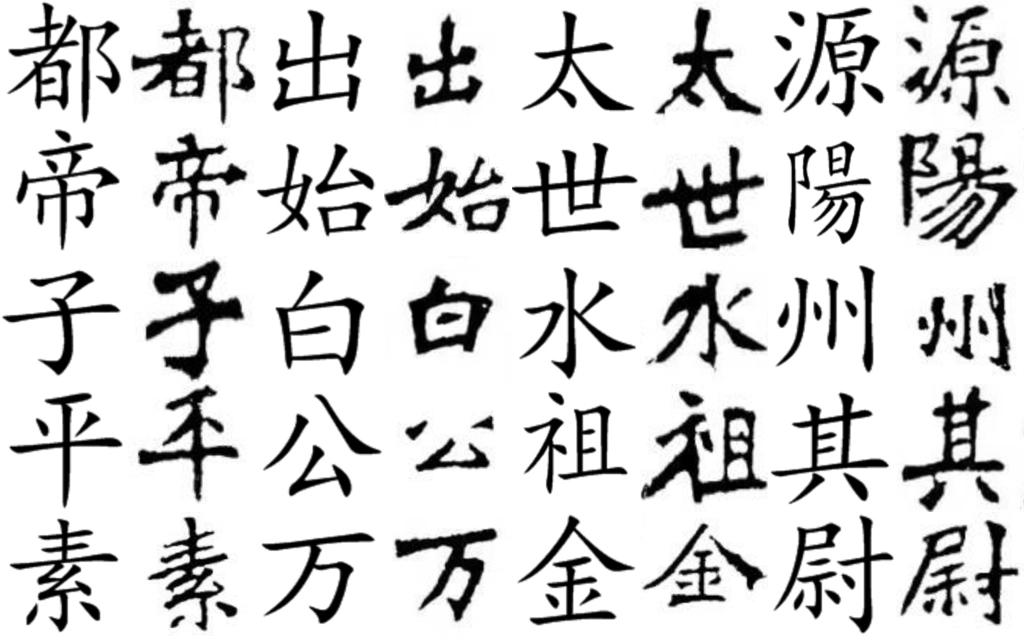}
		\caption{Training Dataset}
		\label{fig:3a}
	\end{subfigure}
	\hfill
	\centering
	\begin{subfigure}{3.8cm}
		\includegraphics[width=3.8cm]{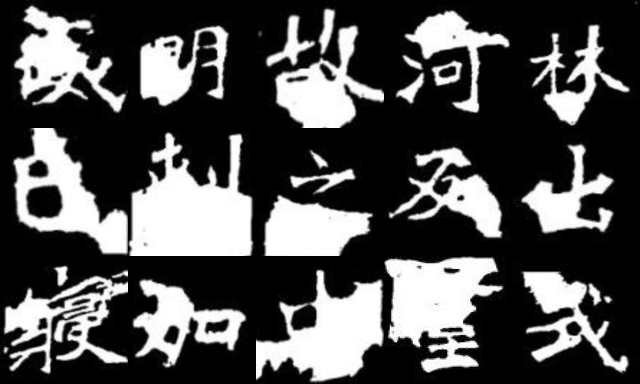}
		\caption{Restoration Dataset}
		\label{fig:3b}
	\end{subfigure}
	\caption{Dataset Overview}
	\label{fig:3}
\end{figure}

To train the model, we first crop the complete characters. The cropping progress is achieved manually. The datasets is illustrated in Figure \ref{fig:3}. 

\par

Since the characters we have cropped are clear, we label the ZhangMenglong style characters through using the same character in another font and joint it horizontally. The training dataset consists of 337 images and the size of the each pair data is $256\times128$ with 3 color channels. The validation dataset consists of 75 images and the image information is as same as the training dataset. In addition there is a reconstruction or restoration dataset, the total number of images in this dataset is 34. The size of images in this dataset is $128\times128$.
\par 
In the training process, we crop the pair data first and get two images that the size are $128\times128\times3$ [standard font image and Zhang Menglong font image]. During the training period, we resize both the standard font characters and the Zhang Menglong font characters in pair data from the size of $128\times128\times3$ to $256\times256\times3$ by using bilinear interpolation. And we convert the image into correspond tensor. The input tensor size for Generator is $256\times256\times3$ and the output tensor size is $256\times256\times3$. For Discriminator1, we concatenate the standard image tensor with the Zhang Menglong font image tensor or with the generated image tensor to get the input tensor. The input tensor size for Discriminator1 is $256\times256\times6$. For Discriminator2, we input the real Zhang Menglong font tensor or the generated font tensor, which size are both  $256\times256\times3$.
\par
For the rest characters in ZhangMenglong Bei, they are slightly or severely damaged, as shown in Fig. \ref{fig:3b}. We collect 34 characters and form a dataset to save these incomplete characters. Each character is exhibited as an image with $128\times 128$ pixels. This dataset is named as the incomplete character dataset. We use the incomplete character dataset to corroborate the model can repair the characters and keep the ZhangMenglong Bei font style.


\section{Proposed Method}
In this section, we will introduce the main architecture of the RubbingGAN in details. The objective of RubbingGAN is to learn a mapping from an original font style source to a specific rubbing font style. We denote the original font KaiTi\_GB2312 as $\{X\}$, and the target ZhangMenglong Bei font the $\{Y\}$. The RubbingGAN is to learn a mapping $G: X \to Y$. 

\subsection{The Architecture of Rubbing GAN}	
\begin{figure}[h]
	\centering
	\caption{RubbingGAN architecture}              
	\includegraphics[width=8cm]{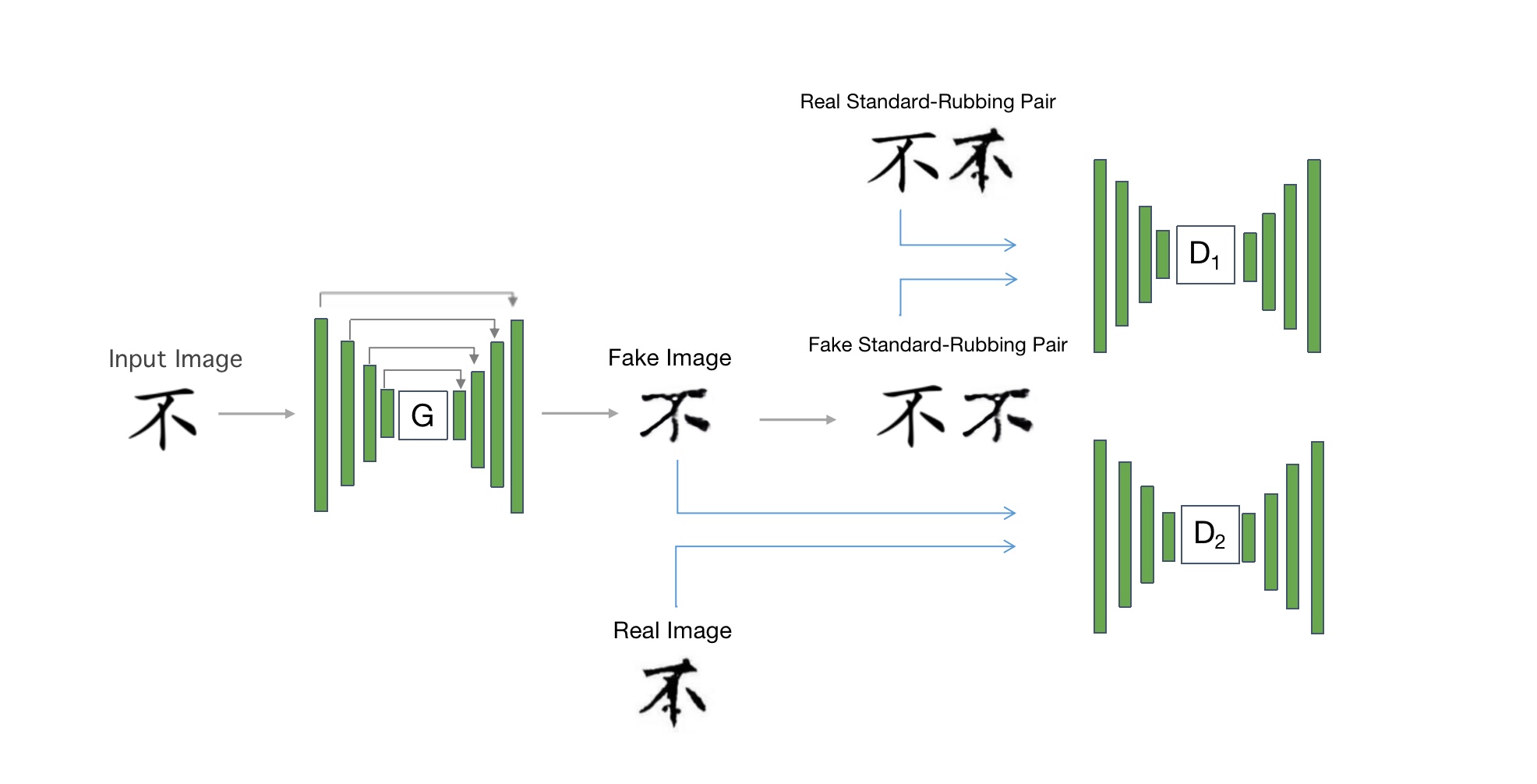}
	\label{fig:4}
\end{figure}

We propose a framework for Rubbing characters restoration that transferred standard font image $x$ into Zhang Menglong font $y$. Our framework is designed based on the pix2pix \cite{ref20} and BEGAN \cite{ref21}. The architecture of our model is shown in Figure \ref{fig:4}.  

The RubbingGAN includes one generator denoted by G and two discriminators, denoted by D1 and D2. 

\begin{figure}[ht]
	\centering              
	\includegraphics[width=8cm]{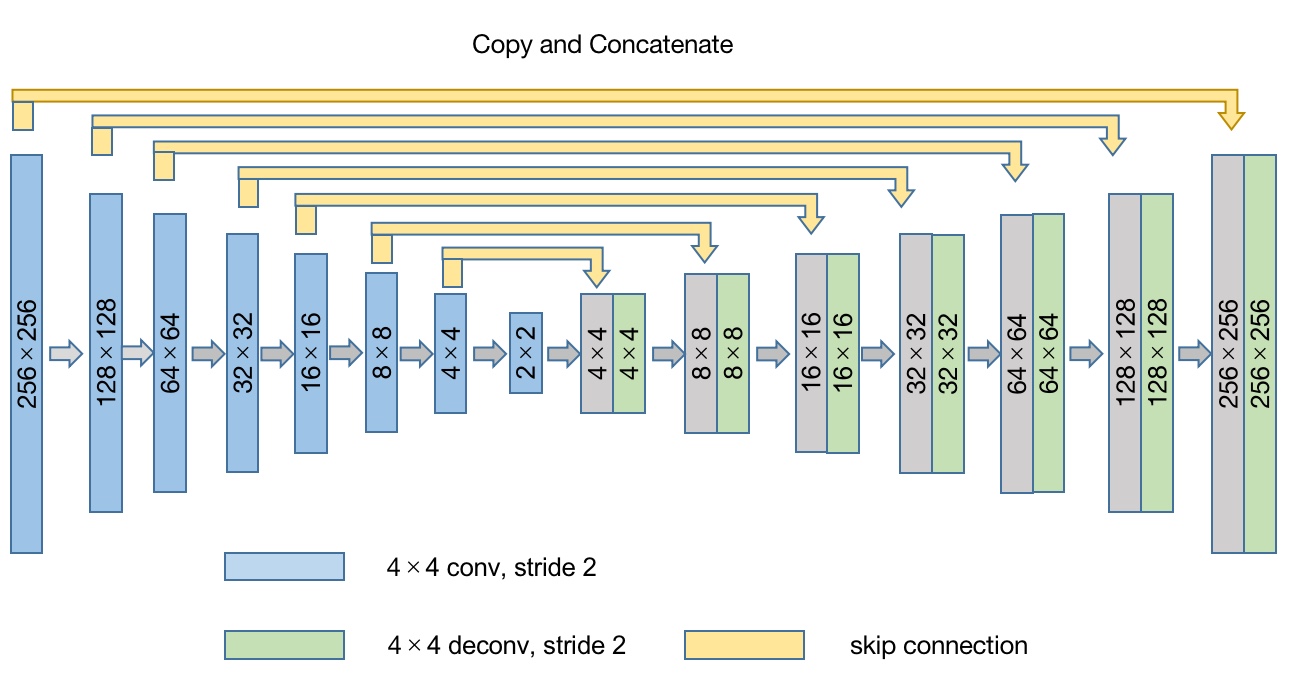}
	\caption{The architecture of U-net. The input size of the image is $256\times256\times3$, the output of the Generator is a Zhangfont style image with size of $256\times256\times3$ }
	\label{fig:5}
\end{figure}

\paragraph{ \textbf{Generator.}} From previous work \cite{ref17} and \cite{ref29}, we adopt "U-Net" structure as our generator which adds skip connections between layers. The Generator structure is shown in Figure \ref{fig:5}. In the rubbing restoration problem, the core problem is to learn the structure changes and the correspond textures. The "U-net" architecture provide high structure semantics with the low-level information.
\par 

\paragraph{ \textbf{Discriminator1.}}
We designed Discriminator1 based on the PatchGAN \cite{ref44}, which encourages GAN to focus both on the high-frequencies and low-frequencies structure features loss. Discriminator1 only penalizes structure at the scale of patches, which can be seen as a type of style/font loss as mentioned in pix2pix.
\par

\begin{figure}[ht]
	\centering              
	\includegraphics[width=8cm]{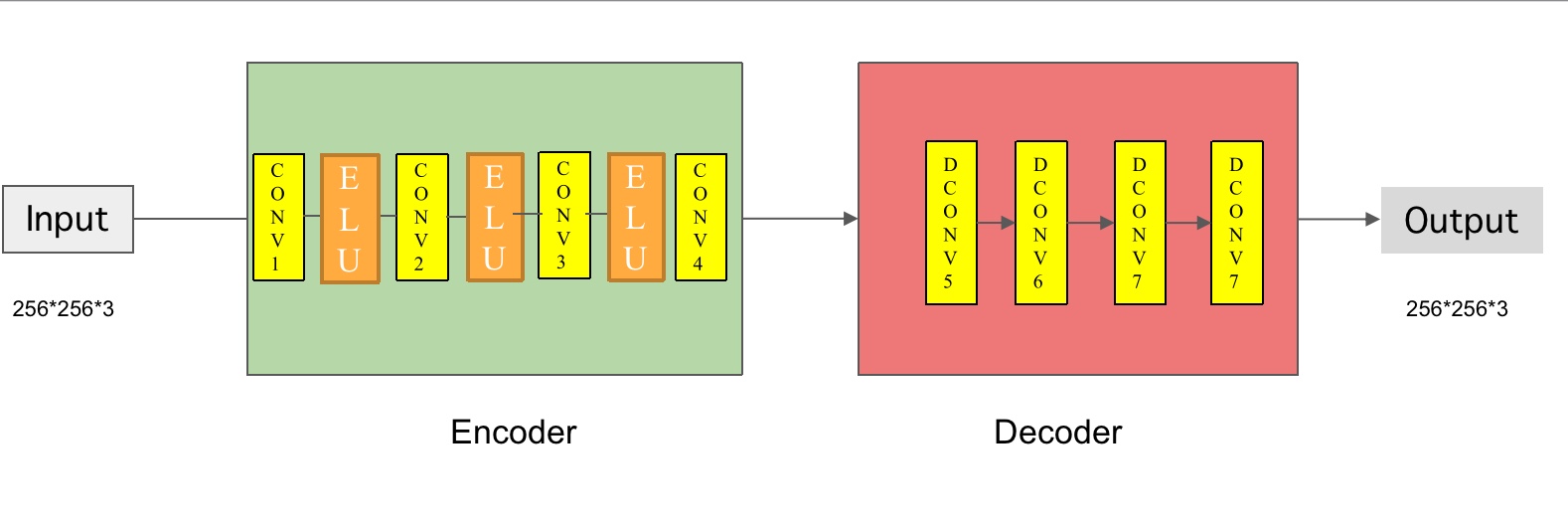}
	\caption{The architecture of Discriminator2.}
	\label{fig:6}
\end{figure}

\paragraph{ \textbf{Discriminator2.}} 
The discriminator2 is an auto encoder-decoder architecture. We designed the structure of Discriminator2 in terms of the BEGAN literature \cite{ref21}. The input size of the image for Discriminator2 is $256\times256\times3$, the output size of the Discriminator is also $256\times256\times3$. The structure of Discriminator2 is really easy, and plays a good effect in image generation. Another advantage of using Discriminator 2 is that in BEGAN researchers propose a method that measured the convergence, which solve the problem of difficulty of training a traditional GAN. Therefore, we would like to use the Discriminator2 to improve generated images quality and also avoid the possibility of mode collapse. The architecture of Discriminator2 is shown in Figure \ref{fig:6}.
\par 

\begin{figure}[h]
	\centering
	\caption{The whole process of our Rubbing restoration}
	\includegraphics[width=8cm]{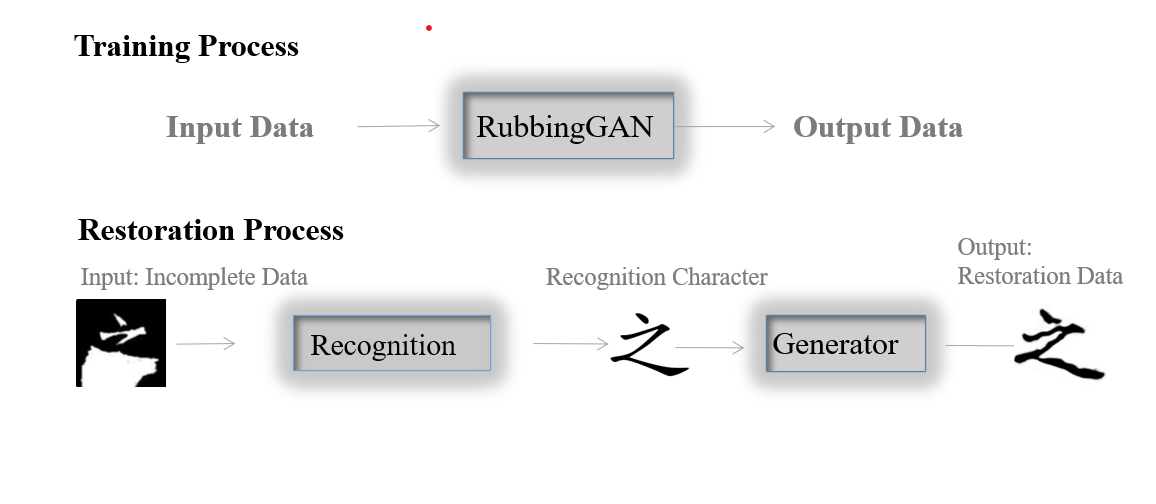}
	\label{fig:7}
\end{figure}

The Figure \ref{fig:7} shows a whole procedure of our rubbing restoration, which recognized the characters first, and then generate the rubbing font characters.
\par

\textbf{Losses}
Our model losses mainly consist of three parts: the adversarial losses from the traditional conditional GAN, the loss from the L1 distance distance, and the loss from the reconstruction.
We use the conditional adversarial loss to generate the realistic image. The loss can be expressed as:
\begin{equation}
	\begin{split}
		\mathcal{L}_{cGAN}(G,D_1) = logD_1(x,y)+log(1-D_1(x,\hat{y}))
	\end{split}
\end{equation}

In order to improve the quality of the generated images, we add the L1 distance into the model. The L1 loss is expressed as: 
\begin{equation}
	\mathcal{L}_{L1}(G)=\mathbb{E}_{x,y}[||{y - G(x)}||] 
\end{equation}	
In order to solve the problem of mode collapses, we use idea from the BEGAN to balance the ability of the generator and discriminator 2.  We use an equilibrium term in RubbingGAN: 
\begin{equation}
	\mathbb{E}\left[\mathcal{L}(G(z))\right]=\gamma\mathbb{E}\left[\mathcal{L}(x)\right]
\end{equation}
where the hype-parameter $\lambda \in [0,1]$ is used to control the output diversity of the Generator. The $\mathcal{L}(\cdot)$ represents the reconstruction loss function for the auto-encoder.
The loss function for the generator and discriminator2 can be expressed as:
\[
\begin{cases}
	\mathcal{L}_{D2}=\mathcal{L}(x)-k_{t}.\mathcal{L}(G(x)) & \textrm{for }\theta_{D2}\\
	\mathcal{L}_{G}=\mathcal{L}(G(x)) & \textrm{for }\theta_{G}\\
	k_{t+1}=k_{t}+\lambda_{k}(\gamma\mathcal{L}(x)-\mathcal{L}(G(x_{G}))) & \textrm{for each training step}
\end{cases}
\]
\par
To sum up, the objective of our model can be expressed as:

\begin{equation}
	{G^*}={arg}\underset{G}{\text{min }}\underset{D_1\ D_2}{\text{max }} \mathcal{L}_{cGAN}(G,D_1) + \lambda \mathcal{L}_{L1}(G) + \mathcal{L}(G,D_2)\\
\end{equation}

where $D_1$ represents the PatchGAN discriminator and the $D_2$ refers to the auto-encoder decoder discriminator.


\section{Experiments}
In this section, we discuss the training procedure and evaluate our model performance on the test dataset. Then, we estimate the font style learning accuracy for the characters based on different criteria. In addition, we investigate the restoration effect by using the criteria.
\par


\subsection{Training Process}
\textbf{Parameters and Optimization}
To optimize RubbingGAN networks, we refer to the steps in pix2pix and BEGAN. We train the Discriminator1 with the Generator alternately. We train the Discriminator1 with Generator first, and then train the Discriminator2 with Generator. The method we trained Discriminator1 and Generator is similar as mentioned in pix2pix. We train the network with the gradient descent step on D1 and then train the Generator with same method. We use the Adam Optimizer which designed from the SGD to train our Generator and D1. The learning rate we adopt is $0.0002$, with the momentum parameters $\beta_1=0.5,\beta_2=0.999$. After training the D1 and Generator, we train the D2 and Generator. For the D2 and Generator, we use the same parameters between the D1 and Generator. 
We analysis the model loss and convergence quantitatively. 
\par

\subsection{Evaluation}
To evaluate the restoration effects, we first measure the font style transfer similarity by using two criteria. Then, we research the restoration effects on our incomplete characters dataset.
\par

\subsection{Font Style Similarity in Testing Set}
In this part, we use the testing dataset to show that the RubbingGAN can learn the font style of ZhangMenglong Bei. We use the paired data to train the RubbingGAN. To evaluate the accuracy of the font style transfer, we adopt the a method called \textbf{Style Similarity Error}. The formula of style similarity error is shown in (3). In detail, the evaluation is achieved based on the complete characters in Zhang Menglong font. We measure the difference between the generated Zhang font with the real Zhang font images. In mathematics, we represent generated font as $X$, the real font as $Y$. And $X$ and $Y$ are two metrics, which dimension are $256\times256$. Our method for evaluation is to calculate square value of every pixel difference in one image. 
\begin{equation}
	\begin{aligned}
		{\frac{1}{m}\sum_{1}^m({X_{i,j}-Y_{i,j}})}^2
	\end{aligned}
\end{equation}
\par

\begin{figure}[ht]
	\centering
	\begin{subfigure}{5cm}               
		\includegraphics[width=5cm]{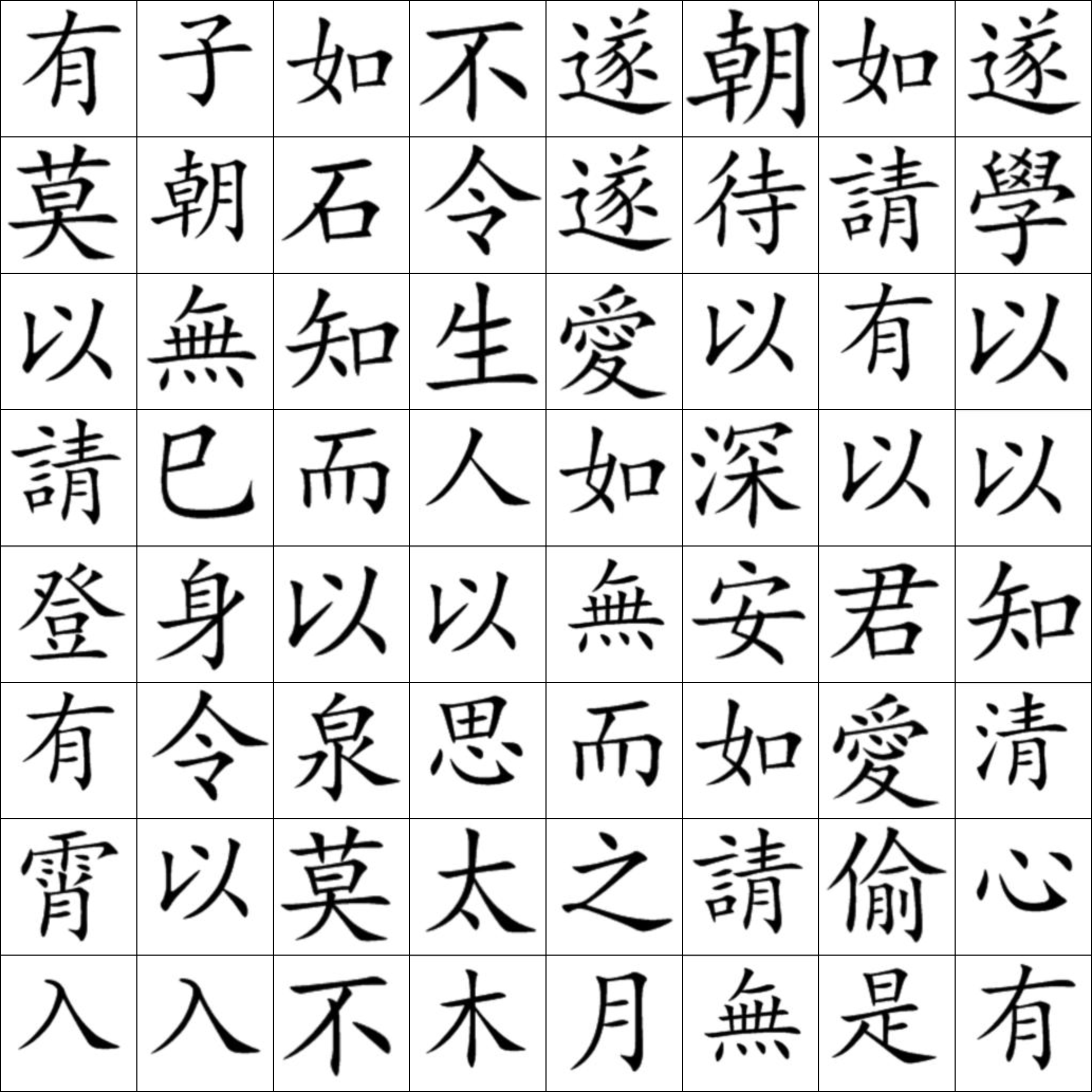}
		\caption{The original font image. }
		\label{fig:8a}
	\end{subfigure}
	\hfill
	\hspace{0in}
	\begin{subfigure}{5cm}
		\includegraphics[width=5cm]{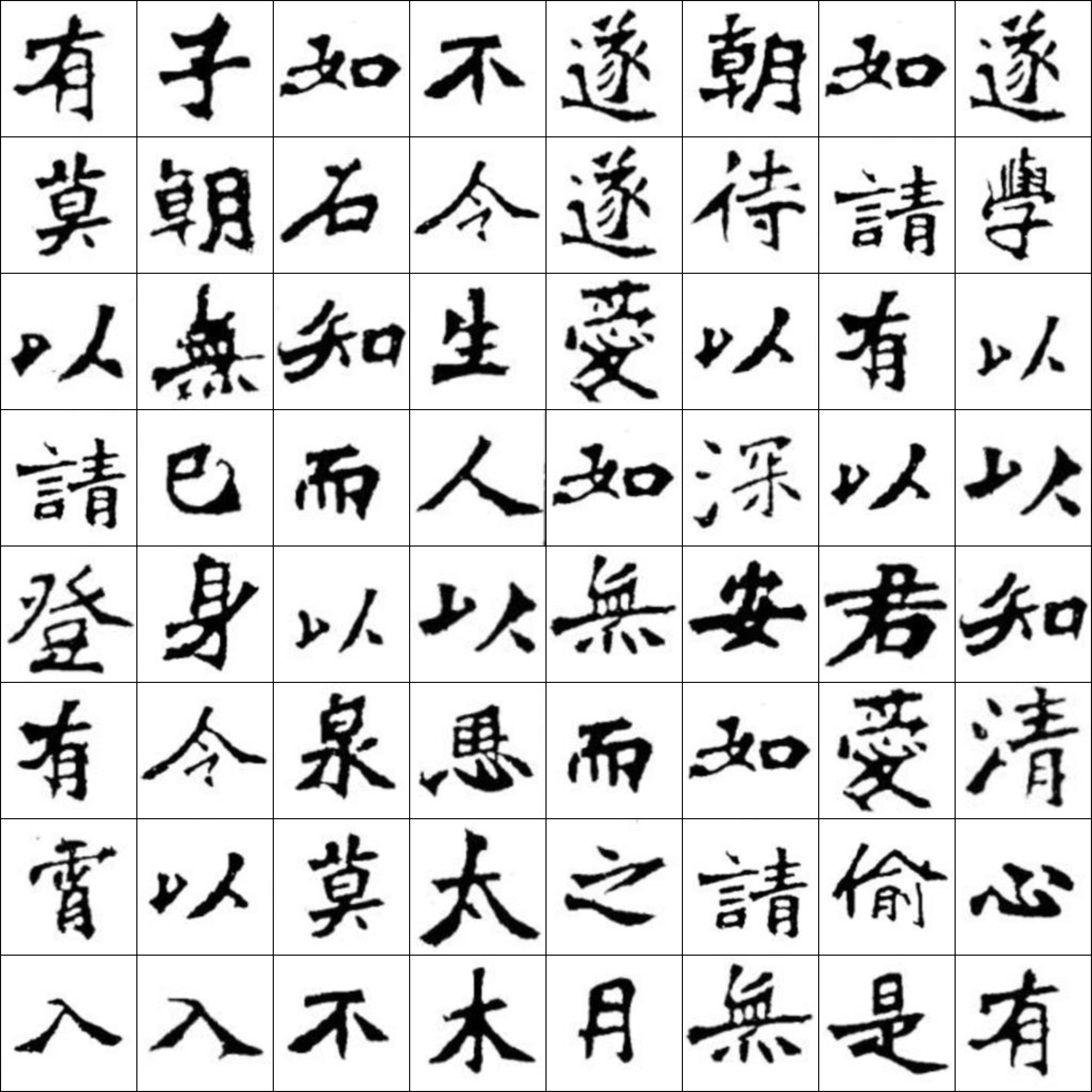}
		\caption{The Zhang font image.}
		\label{fig:8b}
	\end{subfigure}
	\hfill
	\hspace{0in}
	\begin{subfigure}{5cm}
		\includegraphics[width=5cm]{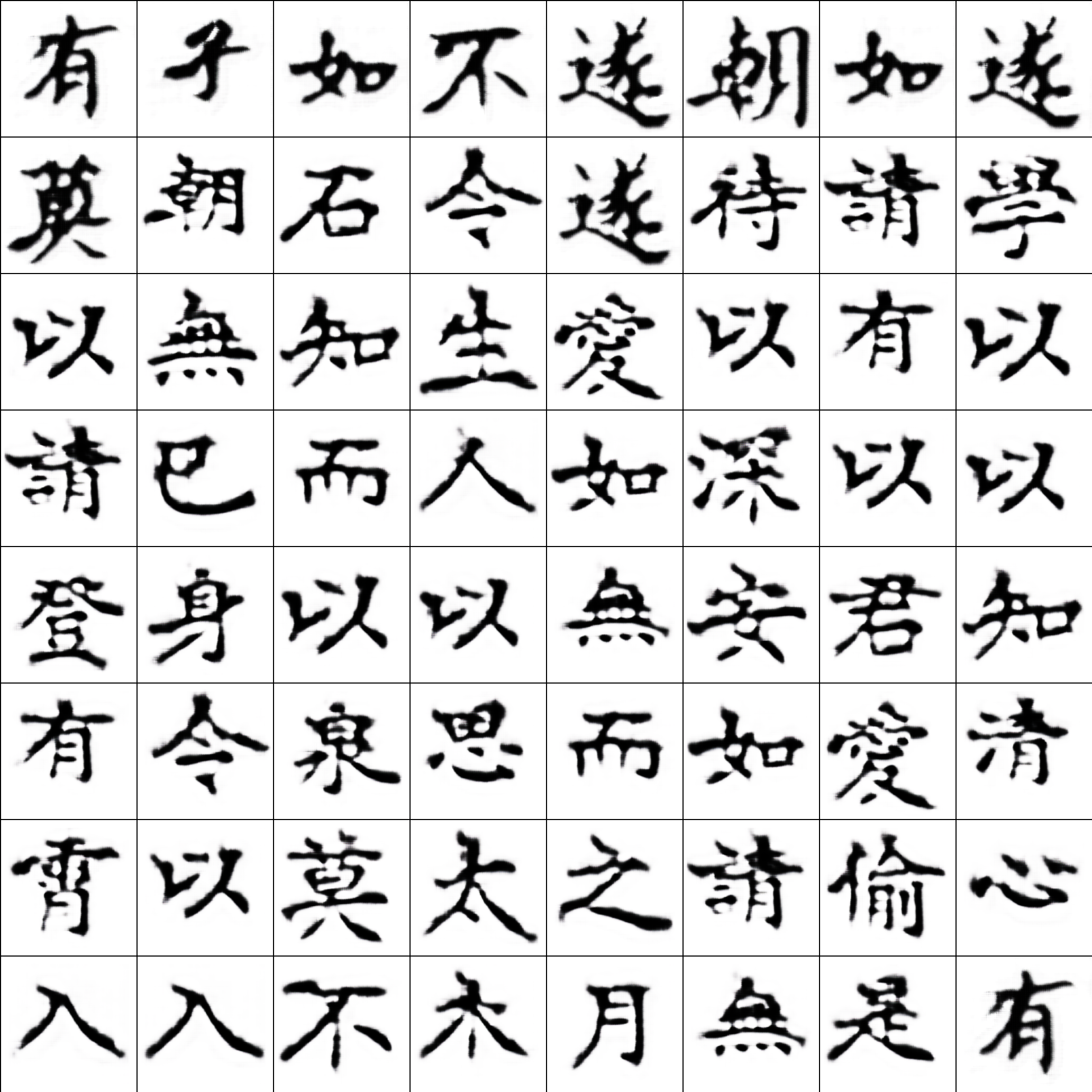}
		\caption{The generated Zhang font image. \small{The characters in first two lines are trained on 9800 iterations. The rest of characters is trained on 4000 iterations.}}
		\label{fig:8c}
	\end{subfigure}
	\caption{Rubbing style transfer result}
	\label{fig:8}
\end{figure}

We use the testing dataset as input for the Generator and validate its performance on rubbing style transfer. The results are shown in Figure \ref{fig:8}.
\par

We use the formula mentioned in the previous section to calculate the similarity between the generated image and original Zhang font image. The result is shown in the Table \ref{t:1}.

\begin{table}[htp]
	\centering
	\caption{\textbf{RubbingGAN} results for Zhang fonts style transfer}
	\label{t:1}
	\begin{threeparttable}
		\begin{tabular}{lccc}
			\hline
			Model&$pix2pix$&$RubbingGAN$&$zi2zi$  \\  \hline
			MSE &0.0006128 &0.0005846 &0.0005028  \\ \hline
		\end{tabular}
	\end{threeparttable}
\end{table}
We find out that RubbingGAN model has a better performance than the pix2pix and a little behind of the zi2zi model.
\par

We also use the FID index to evaluate our model performance. The FID is an index used for evaluating the generating images quality \cite{ref45}. We use the pytorch version to test our model \cite{ref46}. The result is shown in the Table \ref{t:2}.

\begin{table}[hbp]
	\centering
	\caption{\textbf{RubbingGAN} results for image generation quality}
	\label{t:2}
	\begin{threeparttable}
		\begin{tabular}{lccc}
			\hline
			Model&$pix2pix$&$RubbingGAN$&$zi2zi$  \\  \hline
			FID   &218.6381 &204.9481 &270.6007  \\ \hline
		\end{tabular}
	\end{threeparttable}
\end{table}

We find out our model RubbingGAN has the best performance (the lower the better), which suggests our model has a good ability on image generation.
\subsection{Slightly incomplete characters reconstruction}
\begin{figure}[h]
	\centering
	\includegraphics[width=8cm]{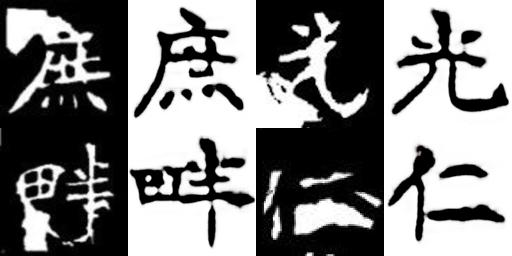}
	\caption{The characters with the black background are the little incomplete characters in Zhang Menglong bei. The characters with the white background are the generated Zhang font characters training on 4000 epochs.}
	\label{fig:9}
\end{figure}

In this experiment, we test the performance on slightly incomplete characters. We first recognize the characters in the slightly incomplete characters dataset. Then, we input the correspond integrated character in standard font through the RubbingGAN and we get the Zhangfont style character. The results are shown in Figure \ref{fig:9}.

\par

\begin{figure}[ht]
	\centering
	\begin{subfigure}{2.4cm}               
		\includegraphics[width=2.4cm]{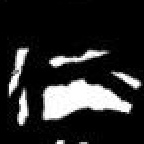}
	\end{subfigure}
	\hfill
	\begin{subfigure}{2.4cm}               
		\includegraphics[width=2.4cm]{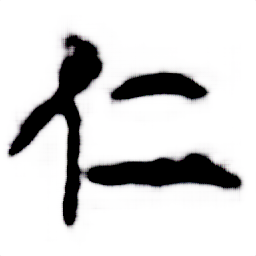}
	\end{subfigure}
	\hfill
	\begin{subfigure}{2.4cm}               
		\includegraphics[width=2.4cm]{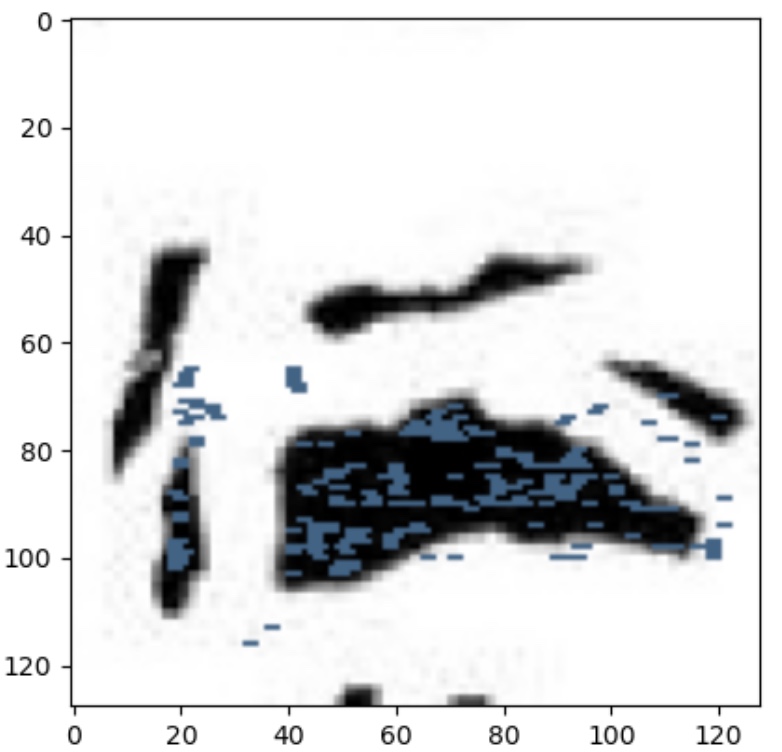}
	\end{subfigure}
	\caption{To calculate the difference between the characters, we first reverse all the pixel value of the incomplete characters in ZhangMenglong Bei. Then if the two image pixel difference is bigger than 253, we mark the pixel with blue color.}
	\label{fig:10}
\end{figure}

We still use the method in experiment A to evaluate our results. However, due to the characters are incomplete, we can't get result quantitatively. However, we can still use the previous formula to consider this problem. If the character are restored successfully, the correspond incomplete parts should be different with the original part. Based on this, we assume that if the difference of pixel value between the generated image and the incomplete image is bigger than 253, then we think this pixel should be in the incomplete part (the value 253 is chose empirically) in the Zhang Menglong Bei. The results are shown in Figure \ref{fig:10}.

\par

\subsection{Severe Incomplete Characters Reconstruction}
\begin{figure}[ht]
	\centering
	\includegraphics[width=8cm]{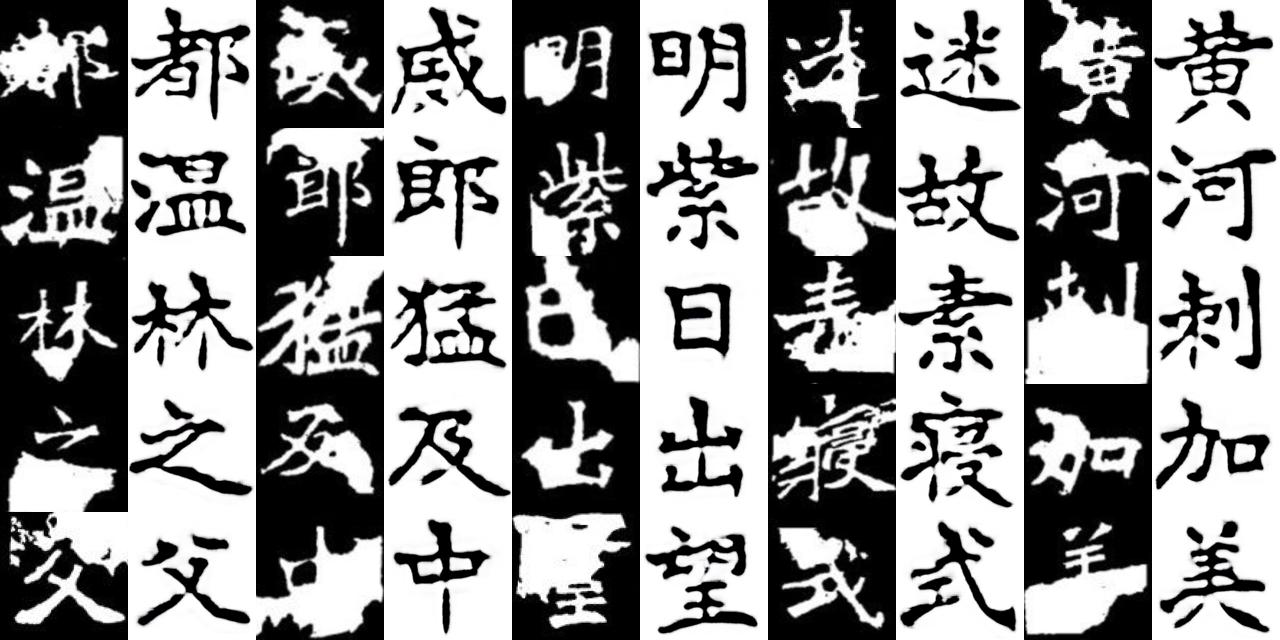}
	\caption{The sever incomplete characters reconstruction}
	\label{fig:11}
\end{figure}

In the previous experiments, we test the model performance on the style transfer, however, the effect of restorations still unknown. Therefore, we use the dataset to generate the Zhang's font image. And we evaluate the restoration effect. The results are shown in Figure \ref{fig:11}.

\par 

\section{Conclusion}
Repairing rubbings is significant for preserving Chinese history and for protecting traditional culture. In this paper, we propose a method named as Rubbing Generative Adversarial Network (RubbingGAN) that can restore the incomplete rubbing characters. Specifically, we collect the characters from a rubbing called Zhang Menglong Bei to the dataset. Based on the dataset we collect, we apply the RubbingGAN to learn the Zhang Menglong Bei font style and restore the characters. In the training process, we prove that our model can converge. The results of our experiment show that generator is good at generating rubbing font characters in the test set and can repair both the slightly and severely incomplete characters.

{\small
\bibliographystyle{ieee}
\bibliography{egbib}
}

\end{document}